\documentclass[a4paper,fleqn]{cas-dc}

\usepackage[numbers]{natbib}

\usepackage{mathtools}
\usepackage{mathrsfs}
\usepackage{amsmath}
\usepackage{array}
\usepackage{bm}
\usepackage{algorithm}
\usepackage[noend]{algpseudocode}
\usepackage{subcaption}
\usepackage[most]{tcolorbox}
\usepackage{placeins}
\usepackage{float} %float barriers
\usepackage{soul}
\graphicspath{ {./images/} }
\newcommand{\bc}{\begin{center}}
\newcommand{\ec}{\end{center}}
\newcommand{\bea}{\begin{eqnarray}}
\newcommand{\eea}{\end{eqnarray}}

\newcommand{\nn}{\nonumber}
\makeatletter
\newcommand*{\rom}[1]{\expandafter\@slowromancap\romannumeral #1@}
\usepackage{marginnote}
\usepackage{soul}

\def\tsc#1{\csdef{#1}{\textsc{\lowercase{#1}}\xspace}}
\tsc{WGM}
\tsc{QE}
\tsc{EP}
\tsc{PMS}
\tsc{BEC}
\tsc{DE}
\begin{document}
\let\WriteBookmarks\relax
\def\floatpagepagefraction{1}
\def\textpagefraction{.001}
\shorttitle{Bias-reduced HER}
\shortauthors{B Manela et~al.}

\title [mode = title]{Bias-Reduced Hindsight Experience Replay with Virtual Goal Prioritization}                      
\tnotemark[1]

\tnotetext[1]{This research was supported in part by the Helmsley Charitable Trust through the Agricultural, Biological and Cognitive Robotics Initiative and by the Marcus Endowment Fund both at Ben-Gurion University of the Negev. This research was supported by the Israel Science Foundation (grant no. 1627/17)}

%\tnotetext[2]{The second title footnote which is a longer text matter
%   to fill through the whole text width and overflow into
%   another line in the footnotes area of the first page.}

%\author[1,3]{B Manela}[type=editor,
%                        auid=000,bioid=1,
%                        prefix=Sir,
%                        role=Researcher,
%                        orcid=0000-0001-7511-2910]
%\cormark[1]
%\fnmark[1]
%\ead{cvr_1@tug.org.in}
%\ead[url]{www.cvr.cc, cvr@sayahna.org}
%
%\credit{Conceptualization of this study, Methodology, Software}
%
%\address[1]{Department of Industrial Engineering and Management, Ben-Gurion University of the Negev, Beer-Sheva, Israel}

\author[1]{B Manela}[]
%\cormark[1]
%\fnmark[1]
\ead{manelab@post.bgu.ac.il}
%\ead[url]{www.cvr.cc, cvr@sayahna.org}

%\credit{Conceptualization of this study, Methodology, Software}

\address[1]{Department of Industrial Engineering and Management, Ben-Gurion University of the Negev, Beer-Sheva, Israel}

\author[1]{A Biess}[orcid=0000-0002-0087-3675]
\cormark[1]
%\fnmark[1]
\ead{abiess@bgu.ac.il}
\ead[url]{www.armin-biess.net}
\cortext[cor1]{Corresponding author}
%\credit{Conceptualization of this study, Methodology, Software, Supervision}

%\author[2,4]{Han Theh Thanh}[style=chinese]
%
%\author[2,3]{CV Rajagopal}[%
%   role=Co-ordinator,
%   suffix=Jr,
%   ]
%\fnmark[2]
%\ead{cvr3@sayahna.org}
%\ead[URL]{www.sayahna.org}
%
%\credit{Data curation, Writing - Original draft preparation}
%
%\address[2]{Sayahna Foundation, Jagathy, Trivandrum 695014, India}
%
%\author%
%[1,3]
%{Rishi T.}
%\cormark[2]
%\fnmark[1,3]
%\ead{rishi@stmdocs.in}
%\ead[URL]{www.stmdocs.in}
%
%\address[3]{STM Document Engineering Pvt Ltd., Mepukada,
%    Malayinkil, Trivandrum 695571, India}
%

%\cortext[cor2]{Principal corresponding author}
%\fntext[fn1]{This is the first author footnote. but is common to third
%  author as well.}
%\fntext[fn2]{Another author footnote, this is a very long footnote and
%  it should be a really long footnote. But this footnote is not yet
%  sufficiently long enough to make two lines of footnote text.}
%
%\nonumnote{This note has no numbers. In this work we demonstrate $a_b$
%  the formation Y\_1 of a new type of polariton on the interface
%  between a cuprous oxide slab and a polystyrene micro-sphere placed
%  on the slab.
%  }

\begin{abstract}
Hindsight Experience Replay (HER) is a multi-goal reinforcement learning algorithm for sparse reward functions. The algorithm treats every failure as a success for an alternative (virtual) goal that has been achieved in the episode. Virtual goals are randomly selected, irrespective of which are most instructive for the agent. In this paper, we present two improvements over the existing HER algorithm. First, we prioritize virtual goals from which the agent will learn more valuable information. We call this property the \textit{instructiveness} of the virtual goal and define it by a heuristic measure, which expresses how well the agent will be able to generalize from that virtual goal to actual goals. Secondly, we reduce existing bias in HER by the removal of misleading samples. To test our algorithms, we built three challenging environments with sparse reward functions. Our empirical results in both environments show vast improvement in the final success rate and sample efficiency when compared to the original HER algorithm. A video showing experimental results is available at \color{blue}{\url{https://youtu.be/xjAiwJiSeLc}}.

%\noindent\texttt{\textbackslash begin{abstract}} \dots 
%\texttt{\textbackslash end{abstract}} and
%\verb+\begin{keyword}+ \verb+...+ \verb+\end{keyword}+ 
%which
%contain the abstract and keywords respectively. 
%
%\noindent Each keyword shall be separated by a \verb+\sep+ command.
\end{abstract}

%\begin{graphicalabstract}
%\includegraphics{figs/grabs.pdf}
%\end{graphicalabstract}

\begin{highlights}
\item Novel technique for virtual goal prioritization provides better value function estimation.
\item Bias-reduction by filtering misleading samples leads to improved learning over vanilla Hindsight Experience Replay (HER).
\item Introduction of new environments for complex manipulation tasks with sparse feedback.
\end{highlights}


\maketitle

\section[Enumerated]{Introduction}
Deep reinforcement learning, the combination of reinforcement learning \cite{sutton2018reinforcement} with deep learning \cite{goodfellow2016deep} has led to many breakthroughs in recent years in generating goal-directed behavior in artificial agents ranging from playing Atari games without prior knowledge and human guidance \cite{mnih2013playing}, to teaching an animated humanoid agent to walk \cite{todorov2012mujoco, lillicrap2015continuous, schulman2017proximal}, and defeating the best GO player in the world \cite{silver2018general}, just to name a few. All reinforcement learning problems are based on the reward hypothesis, stating that any goal-directed task can be formulated in terms of a reward function. However, the engineering of such a reward function is often challenging. The difficulties in shaping suitable reward functions limit the application of reinforcement learning to real-world tasks, for example, in robotics \cite{kober2013reinforcement}. One way to overcome the problem of reward shaping has been presented in Hindsight Experience Replay (HER) \cite{andrychowicz2017hindsight}, which uses sparse reward signals to indicate whether a task has been completed or not. The algorithm uses failures to learn how to achieve alternative goals that have been achieved in the episode and uses the latter to generalize to actual goals. HER selects these virtual goals randomly in every episode.
	
Our contributions in this paper consists of two improvements over the original HER algorithm for the selection of virtual goals. First, we argue that the learning process will be more efficient if the algorithm will take into account that some virtual goals may be more instructive than others and will prioritize them accordingly. Towards this objective, we present a heuristic measure, which quantifies the instructiveness of each possible virtual goal. We call this method an \textit{Instructional-Based Strategy} (IBS). Our strategy for selecting virtual goals will be applicable to environments for which the initial state distribution is \textit{not} within the goal distribution. These conditions are given, for example, in robotic manipulators, where objects need to be moved between two different locations in a workspace. 
Our second contribution consists of the removal of misleading virtual goals by providing a filtered version of HER  (\textit{Filtered HER}). HER assumes that the achieved goals are the outcome of actions. However, this is not always the case. Consider, for example, a soccer player that misses the ball in a penalty kick. Here, the ball is not affected by the action, and thus, the achieved goal is the outcome of the initial state rather than the action. These samples will induce bias and may hinder learning. A filtering process can be applied to remove these misleading samples.
To the best of our knowledge, we present the first method to identify and remove biased transitions produced by HER. Finally, we introduce three challenging manipulation tasks to benchmark the performance of our algorithms.

\section[Enumerated]{Background}
In this section we provide the background information for reinforcement learning and hindsight experience replay.
\subsection[Enumerated]{Reinforcement Learning}
Reinforcement learning consists of an agent learning how to solve a task via interaction with an environment \cite{sutton2018reinforcement}. We assume that the environment is fully observable and defined by a set of states $s \in \mathcal{S}$, set of actions $a \in \mathcal{A}(s)$, initial state distribution $P(s_0)$, reward function $r: \mathcal{S} \times \mathcal{A} \rightarrow \mathbb{R}$ and a discount factor $\gamma \in [0,1]$. These components define a \textit{Markov Decision Process} (MDP). The decisions of the agent are described by a policy $\pi$ that maps states $s$ to actions $a$. At the beginning of each episode, an initial state $s_0$ is sampled from the distribution $p(s_0)$. At every timestep, the agent chooses an action $a_t$ using policy $\pi(s_t)$, performs this action and receives a reward and the next state. The episode is terminated when the agent reaches a terminal state or exceeds the maximum number of timesteps. The agent's goal is to find the policy that maximizes expected return, i.e., cumulative future discounted reward $R_t=\sum_{i=t}^{\infty}\gamma^{i-t}r_i$.

\subsection[Enumerated]{Deep Deterministic Policy Gradient (DDPG)}
\hspace*{-0.2cm}RL algorithms can be implemented using temporal-differ-\\ence learning, policy gradient or a combination of both in actor-critic methods. One of the most prominent actor-critic algorithms for continuous state- and action spaces is \textit{Deep Deterministic Policy Gradient} (DDPG) \cite{lillicrap2015continuous}. 

The DDPG architecture consists of two neural networks: an \textit{actor}, which takes the state $s_t$ as input and outputs the chosen action $a_t$, and a \textit{critic}, which approximates the $Q$-function, $Q(s_t, a_t)$, for the chosen action. The critic network is trained using temporal-difference with loss function ${\cal{L}}_c = \frac{1}{N}\sum_{i}(y_i - Q(s_i,\pi(s_i)))^2$ with action $a=\pi(s_i)$ generated by the actor and target $y_i=r_i+\gamma Q'(s_{i+1},\pi'(s_{i+1}))$, where $Q', \pi'$ are provided by the target networks. The actor is trained using gradient descent on the loss\\
${\cal{L}}_a=-\mathbb{E}_{\rho(s)}Q(s, \pi(s))$.

\subsection{Hindsight Experience Replay (HER)}
DDPG can be extended to multi-goal tasks using \textit{Universal Value Function Approximators} (UVFA) \cite{schaul2015universal}. The key idea behind UVFA is to augment action-value functions and policies by goal states, and thus, every transition contains also the desired goal. This enables generalization not only over states but also over goals when using neural networks as function approximators. 
In multi-goal tasks with \textit{sparse} rewards, it is challenging to learn the task and achieve any progress. HER addresses this problem by taking failure as a success to an alternative (or \textit{virtual}) goal. HER applies UVFA and includes additional transitions with virtual goals - alternative goals that the agent achieved in the trajectory. Thus, the agent can learn from failures through generalization to actual goals. It has been demonstrated that HER significantly improves the performances in various challenging simulated robotic environments \cite{andrychowicz2017hindsight}. Our algorithms - like HER - can only be used in combination with off-policy algorithms since replacing the goal with a virtual-goal may lead to a different policy than the one which generated the trajectory.

\section{Method} \label{sec:Method}

In this section we present two algorithms: a novel strategy for virtual goal selection, which we call an \textit{Instructional-Based Strategy} (IBS) and a \textit{Filtered-HER} algorithm, which is a method to remove misleading samples in HER.

\subsection{Instructional-Based Strategy (IBS)} \label{subsec:IBS}
In this section we present the \textit{Instructional-Based Strategy} for virtual goal selection and provide pseudo-code for the implementation.
\subsubsection{Motivation}
HER is based on generalizing from previous failures to the desired target. In this paper, we address the question of how these failures should be taken into consideration in the learning process. Is every failure equally instructive as any other as has been proposed by the original HER algorithm? To analyze this question, 
consider the following soccer scenario: a player takes two penalty kicks. In a first kick, the goal was missed by a small distance to the right, whereas in a second kick the goal was missed by far to the left. The question arises which of these experiences is more instructive to the soccer player for learning the task of hitting the goal. It seems that nearly missing the goal is more instructive for achieving the goal. However, it might be that the player has experienced many kicks of the first type whereas none of the second. In this case, the latter kick may be more instructive for learning the task. On this \textit{Instructional-Based Strategy} (IBS), we based our heuristic approach towards virtual goal prioritization. 

\subsubsection{Definitions}
Prioritizing virtual goals is guided by three heuristic principles, which define (i) what the agent needs to learn (ii) what the agent can learn from an individual virtual goal and (iii) what is unknown to the agent. These three principles define the \textit{instructivness} of the virtual goal.\\

\paragraph{What the agent needs to learn: \label{par:need-to-learn}}The task of the agent is to learn the behavior which achieves the actual goals. The goals are described by a goal distribution density $g(\bm{x})$,  where \\
${\mbox{Pr}}(\bm{x} \in {{G}})= \int_{{G}}g(\bm{x})d\bm{x}$ for any measurable set $G \in \mathbb{R}^n$. {The goal distribution density is an input to the algorithm and can be described in most cases by a uniform distribution over the distribution's support ${\cal{G}}$, i.e.,}
\bea \label{eq:uniform}
\label{uniform} % uniform distribution
g(\bm{x}) = \begin{cases}
	const\,, & \mbox{if} \; \bm{x} \in {\cal{G}} \label{1}\\
	0\,, & \mbox{otherwise}\,.
\end{cases}
\eea 
The goals can also be chosen from a non-uniform or non-stationary distribution $g(\bm{x})$ as we discuss in Appendix \ref{appndx:A}.

\paragraph{What can be learned from a virtual goal: \label{par:can-learn}} The selection of a virtual goal $\tilde{\bm{g}}$ (coordinates of a specific goal) teaches the agent of how to reach this goal ($\tilde{\bm{g}}$) as well as other goals that are in the near surrounding of  $\tilde{\bm{g}}$. The latter is due to the generalization capabilities of the underlying neural networks \cite{schaul2015universal,zhang2016understanding}. To approximate the relevance of the virtual goal $\tilde{\bm{g}}$ to other neighboring goals $\tilde{\bm{g}}'$, we use a Gaussian radial basis function (RBF) kernel
\bea
k(\tilde{\bm{g}},\tilde{\bm{g}}'|\sigma) &=& \exp\bigg(-\frac{{\left\lVert \tilde{\bm{g}}-\tilde{\bm{g}}' \right\rVert}^2}{2\sigma^2}\bigg)\,, \label{2}
\eea
	
where $\left\lVert \cdot \right\rVert$ is the $L_2$-norm.
Thus, the relevance of virtual goal $\tilde{\bm{g}}$ to point $\tilde{\bm{g}}'$ is defined by the Mahalanobis distance. We set in (\ref{2}) $\bm{\Sigma} = \sigma^2 \bm{I}$, where the variance $\sigma^2$ is a hyperparameter. Using kernel regression we score virtual goals using the goal distribution
\bea \label{eq:value}
\mu({\tilde{\bm{g}}}|{\sigma})= \int_{\tilde{\bm{g}}' \in \mathbb{R}^n}
k(\tilde{\bm{g}},\tilde{\bm{g}}' | \sigma) g(\tilde{\bm{g}}') d\tilde{\bm{g}}'\,. \label{3}
\eea
For a uniform goal distribution equation (\ref{eq:value}) simplifies to
\bea \label{eq:simplify_weight}
\mu({\tilde{\bm{g}}}|\bm{\sigma})= const \cdot \int_{\tilde{\bm{g}}' \in {\cal{G}}} k(\tilde{\bm{g}},\tilde{\bm{g}}' | \sigma)d\tilde{\bm{g}}'\,.
\eea
Thus, virtual goals that are located near the distribution's edges receive a lower score. For this reason this strategy will not work for environments where the initial state distributions is within the goal distribution, because initial states will get the highest scores. Scores can be turned into a probability distribution over the possible virtual goals by normalization, resulting in the target distribution $q^\ast$ of virtual goals
\bea \label{eq:optVG}
q^\ast({\tilde{\bm{g}}}|\bm{\sigma}) = \frac{\mu({\tilde{\bm{g}}}|\bm{\sigma})}{\int_{\tilde{\bm{g}} \in \tilde{\cal{G}}} \mu({\tilde{\bm{g}}}|\bm{\sigma})d \tilde{\bm{g}}}\,,
\eea
where $\tilde{\cal{G}}$ denotes the range of all virtual goals. 

\paragraph{What is unknown to the agent: \label{par:unknown}} The agent's current knowledge about the goal distribution is represented by the proposal distribution $q(\tilde{\bm{g}})$ of virtual goals and is initialized with zero. The mismatch between the proposal and target distribution is calculated using the clipped local difference, which is bounded from below by zero, and given by
\bea \label{eq:weight2}
w(\tilde{\bm{g}})= \max[q^*(\tilde{\bm{g}})-q(\tilde{\bm{g}}),0\big]\,.
\eea
Normalization leads to the probability used for prioritization
\bea \label{eq:priority}
p(\tilde{\bm{g}}) = \frac{w(\tilde{\bm{g}})} {\sum_{\tilde{\bm{g}}'\in \tilde{\cal{G}}} w(\tilde{\bm{g}}') }\,, \forall \, \tilde{\bm{g}} \in \tilde{\cal{G}} \,,
\eea
where $\tilde{{\cal{G}}}$ denotes the set of virtual goals. In practice, we find it useful to clip the weights to some small value (we used 0.002) instead of zero, so all virtual goals have some probability of getting sampled. This trick makes learning more stable. 

\subsubsection{Implementation} \label{sec:implementation}
For the implementation of the algorithm we discretize the range of virtual goals $\cal{\tilde{G}} = \cal{S}$ into $M\times N$ grid cells and approximate the target and proposal distributions of virtual goals over the grid cells as

\bea \label{eq:discrete_optVG}
q^\ast(\tilde{\bm{g}}=(i,j) | \bm{\sigma}) &=& \frac{\mu((i,j)|\bm{\sigma})}{ \sum_{i,j=1}^{M,N}\mu((i,j)|\bm{\sigma})}\,,
\eea
\bea \label{eq:discrete_VG}
q(\tilde{\bm{g}}=(i,j)) &=& \frac{1}{|R|}\sum_{\tilde{\bm{g}} \in R} [ \tilde{\bm{g}} \in \mbox{cell}(i,j)]
\eea
for $i=1,\dots,M\,, j=1,\dots, N$, where $(i,j)$ denotes the center of the grid cells, $[\cdot]$ is the indicator function and $R$ the replay buffer of virtual goals with size $|R|$.
To stabilize the learning, we initialize the hyper-parameter $\sigma^2$ to a high value ($2$) and gradually decrease it to its final value ($0.2$) by a decreasing factor of $0.9$ after every $50$ training cycles. The weight of the virtual goal $\tilde{\bm{g}}$ is the weight of its bin
\bea \label{eq:discrete_weight}
w(\tilde{\bm{g}}) &=&
\max\big[q^\ast(bin(\tilde{\bm{g}})|\bm{\sigma})-{q(bin(\tilde{\bm{g}}))} \,,\, 0\big]\,,
\eea
and the prioritization probability is defined as in equation (\ref{eq:priority}). See Alg.\ref{alg:IBS_HER} for a pseudo-code of the algorithm.

\begin{algorithm}
	\caption{Instructional-Based HER} \label{alg:IBS_HER}
	\begin{algorithmic}[1]
		\Require
		\Statex \textbullet~ an off-policy RL algorithm $\mathbb{A}$,\Comment e.g. DQN, DDPG 
		\Statex \textbullet~ a reward function: $\mathcal{S}\times \mathcal{A}\times \mathcal{G} \rightarrow \mathcal{R}$,\Comment e.g. $r(s,a,g)=-1 \text{ if fail, } 0 \text{ if success}$
		\Statex \textbullet~ a goal distribution density $g(\bm{x})$
		\Statex \textbullet~ std $\sigma$ for the target distribution $q^\ast$
		\Statex
		\Statex \textbf{Note:} || denotes concatenation of vectors
		
		\State Initialize $\mathbb{A}$
		\State Initialize replay buffer $R, \quad |R| \leftarrow 0$
		\State Initialize proposal distribution $q$ \Comment $q_{ij}=0 \quad\forall i,j \in [1\ldots M],[1\ldots N]$
		\State Calculate $q^*$ \Comment Using equation (\ref{eq:discrete_optVG})
		\While{True}
		\For{$Episode \leftarrow 1, M$}
		\State Sample a goal $g$ and an initial state $s_0$
		\For{$t \leftarrow 0, T-1$}
		\State Sample an action $a_t \leftarrow \pi(s_t||g)$
		\State Execute $a_t$ and observe a new state $s_{t+1}$
		\EndFor
		
		\For{$t \leftarrow 0, T-1$} \Comment IBS
		\State Calculate the priority $p(\widetilde{g}_t)$ via equation (\ref{eq:priority})
		
		\EndFor
		
		\For{$t \leftarrow 0, T-1$}
		\State $r_t := r(s_t, a_t, g)$
		\State Store the transition $(s_t||g, a_t, r_t, s_{t+1}||g)$ in $R$ \Comment standard experience replay
		
		\State Sample a set of virtual goals $\tilde{G}$ for replay from the future state based on priority $p^*(\widetilde{g})$ 
		\For{$\widetilde{g} \in \tilde{G}$}
		\State $\widetilde{r} = r(s_t, a_t, \widetilde{r})$
		\State Store $(s_t||\widetilde{g}, a_t, \widetilde{r}, s_{t+1}||\widetilde{g})$ in $R$\Comment HER
		\State $|R| \leftarrow |R| + 1$
		\State Update $q$
		\EndFor
		\EndFor
		\EndFor
		
		\For{$t \leftarrow 1, N$}
		\State Sample a minibatch $B$ from the replay buffer $R$
		\State Perform one step of optimization using $\mathbb{A}$ and minibatch $B$
		\EndFor
		\EndWhile
		
	\end{algorithmic}
\end{algorithm}

\subsubsection{Comparison to Reward Shaping}
At a first look, IBS may seem similar to reward shaping, which we aimed to avoid from the beginning. However, these two concepts are fundamentally different. While the objective of reward shaping is to find a feedback signal that describes how close the agent is to task goal completion \cite{dewey2014reinforcement}, IBS relies on the same assumption as in UVFA, namely, that one goal can be generalized to another by using the generalization capabilities of neural networks. Under this assumption, IBS can be applied to \textit{any} task with no further modifications.

\subsection{Filtered-HER} \label{ss:FilteredHER}
In this section we identify a problem in the existing HER algorithm and provide a second improvement over the vanilla-HER algorithm by the removal of misleading samples, resulting in the \textit{Filtered-HER} algorithm.

\subsubsection{Bias in HER} \label{ss:bias in HER}
In this section, we discuss a fundamental problem within the original HER algorithm. A more mathematical explanation is provided in Appendix \ref{appndx:B}. As mentioned in \cite{plappert2018multi}, HER may induce bias in the learning process. Using the \textit{achieved-goal} as a virtual goal may lead in some cases to situations in which the agent performs poorly, even though it repeatedly receiving rewards indicating that it should continue to act in this way. Consider the bit-flipping environment, introduced in \cite{andrychowicz2017hindsight} with few modifications. In the bit-flipping environment, the state- and action spaces are $S = \{0, 1\}^n$ and $A = \{0,1,...,n-1\}$, respectively, for some length $n$. Executing the $i$-th action flips the $i$-th bit of the state. The initial and target states are sampled uniformly at the beginning of each episode. Each step has a cost of $-1$. To illustrate HER's problem, we add a new action, which has no effect on the bits and then terminates the game. Although this action is useless, the agent may think otherwise. As the state remains the same, this state's virtual goal will always be the state itself; thus, the virtual reward of this action will always be positive (zero). As a result, the agent may assume this action is desired. This scenario frequently happens in manipulation tasks, such as in the \textit{Push} task of OpenAI Gym. In this environment, a manipulator needs to push a box to the desired location. If the manipulator does not touch the box, the achieved-goal (i.e., the box position) will not change. Hence, when virtual goals for experience replay are sampled, they all will be the same and identical to all the achieved-goals, resulting in misleading positive virtual rewards. This drawback of HER is similar to the role of terminal states in bootstrapping, in which the values of all states are gradually updated except for terminal states. Terminal states are, by definition, states for which the achieved goal is identical to the desired goal. However, no actions are assigned to terminal states in bootstrapping, nor is any next-state observed (i.e., a tuple $S,A,R,S^\prime$), because assigning actions to terminal states will disturb the learning process.  

\subsubsection{Method}
To resolve this problem, we apply a filter to remove misleading samples. Before storing the virtual sample in the replay buffer, the filter checks if the virtual goal has been already achieved in the current state. If so, the sample will be deleted, and the next virtual goal will be generated. See Alg. \ref{alg:Filtered-HER} for the pseudo-code of the \textit{Filtered-HER} algorithm.

\begin{algorithm}
	\caption{Filtered-HER} \label{alg:Filtered-HER}
	\begin{algorithmic}[1]
		\Require
		\Statex \textbullet~ an off-policy RL algorithm $\mathbb{A}$,\Comment e.g. DQN, DDPG 
		\Statex \textbullet~ a reward function: $\mathcal{S}\times \mathcal{A}\times \mathcal{G} \rightarrow \mathcal{R}$,\Comment e.g. $r(s,a,g)=-1 \text{ if fail, } 0 \text{ if success}$
		\Statex
		\Statex \textbf{Note:} || denotes concatenation of vectors
		
		\State Initialize $\mathbb{A}$
		\State Initialize replay buffer $R$
		
		\While{True}
		\For{$Episode \leftarrow 1, M$}
		\State Sample a goal $g$ and an initial state $s_0$.
		\For{$t \leftarrow 0, T-1$}
		\State Sample an action $a_t \leftarrow \pi(s_t||g)$
		\State Execute $a_t$ and observe a new state $s_{t+1}$
		\EndFor
		
		\For{$t \leftarrow 0, T-1$}
		\State $r_t := r(s_t, a_t, g)$
		\State Store the transition $(s_t||g, a_t, r_t, s_{t+1}||g)$ in $R$ \Comment standard experience replay
		
		\State Sample a set of virtual goals $\tilde{G}$ for replay $\tilde{G} := \mathbb{S}(\text{\textbf{current episode}})$
		\For {$\widetilde{g} \in \tilde{G}$}
		\State $\widetilde{r} = r(s_t, a_t, \widetilde{g})$
		\If{$r(s_{t-1}, a_{t-1}, \widetilde{g})=0$}
		\State Skip transition \Comment \textbf{Filtered-HER}
		\EndIf
		\State Store $(s_t||\widetilde{g}, a_t, \widetilde{r}, s_{t+1}||\widetilde{g})$ in $R$\Comment HER
		\EndFor
		\EndFor
		\EndFor
		
		\For{$t \leftarrow 1, N$}
		\State Sample a minibatch $B$ from the replay buffer $R$
		\State Perform one step of optimization using $\mathbb{A}$ and minibatch $B$
		\EndFor
		\EndWhile
		
	\end{algorithmic}
\end{algorithm} 

\section{Experiments}
Our algorithms were implemented and validated in three ball-throwing environments with different levels of complexity, which we describe next.
\subsection{Environments} \label{environments}
\begin{figure*}
	\centering
	\begin{subfigure}[b]{.3\textwidth}
		\centering
		\tcbox{\includegraphics[width=3cm]{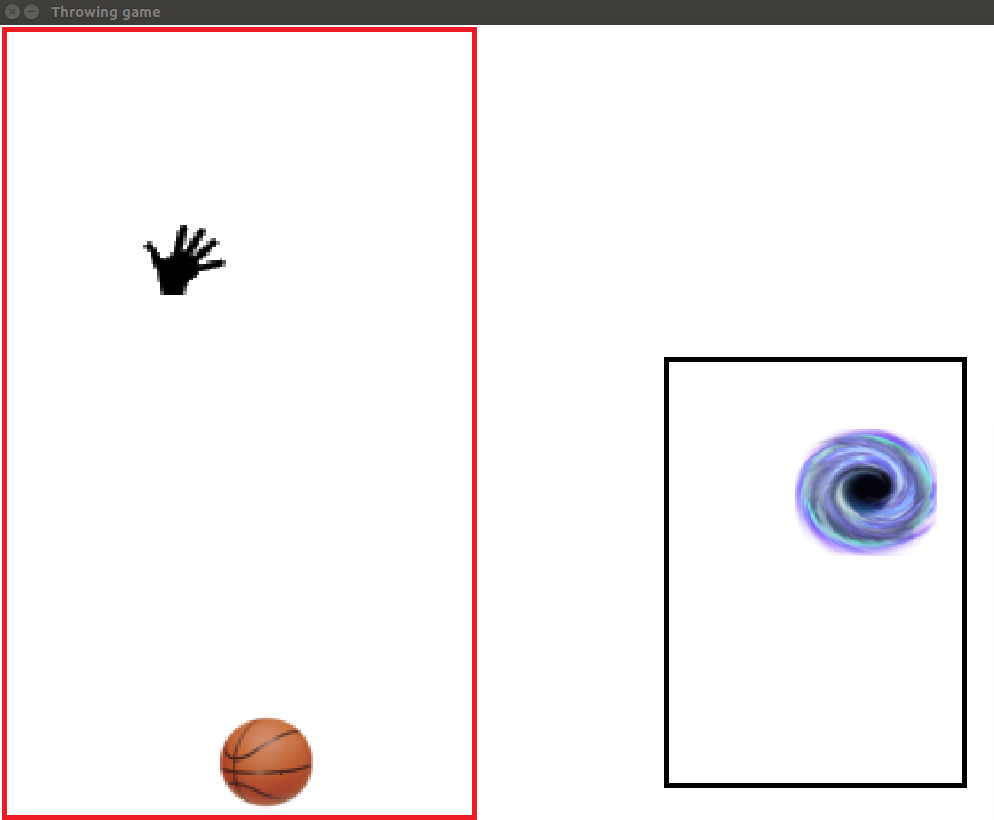}}
		\caption{Hand Environment}
		\label{fig:hand-game}
	\end{subfigure}%
	\begin{subfigure}[b]{.3\textwidth}
		\centering
		\tcbox{\includegraphics[width=3cm]{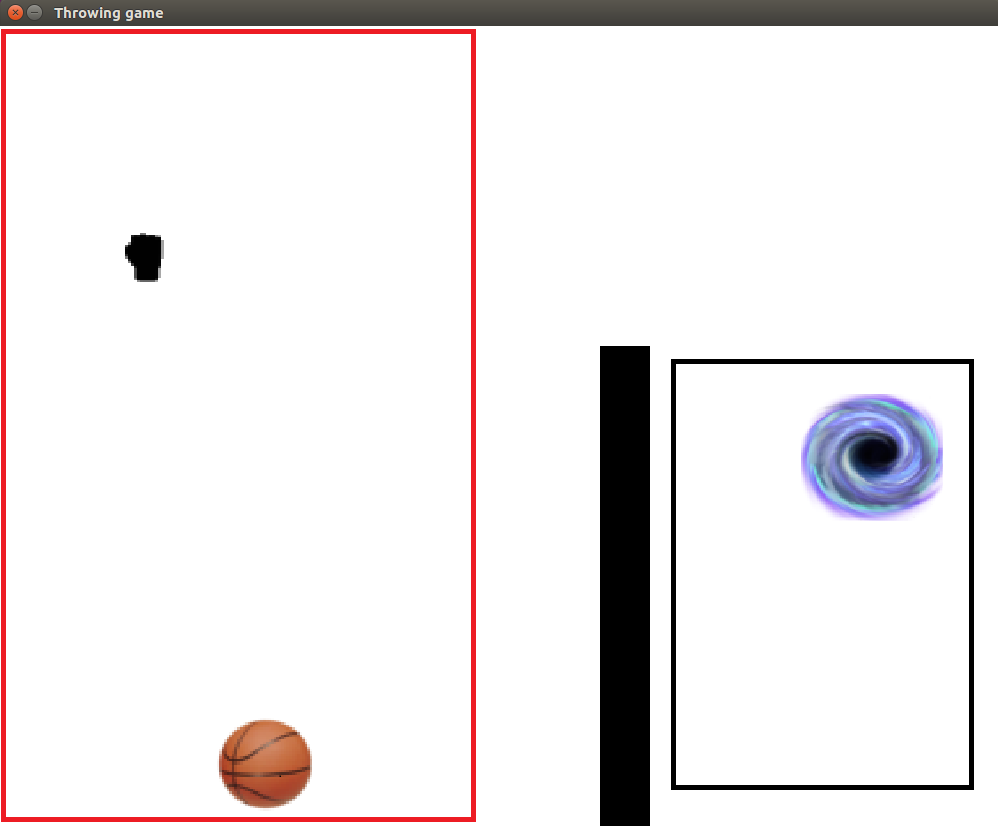}}
		\caption{Hand-Wall Environment}
		\label{fig:hand-wall-game}
	\end{subfigure}
	\begin{subfigure}[b]{.3\textwidth}
		\centering
		\tcbox{\includegraphics[width=3cm]{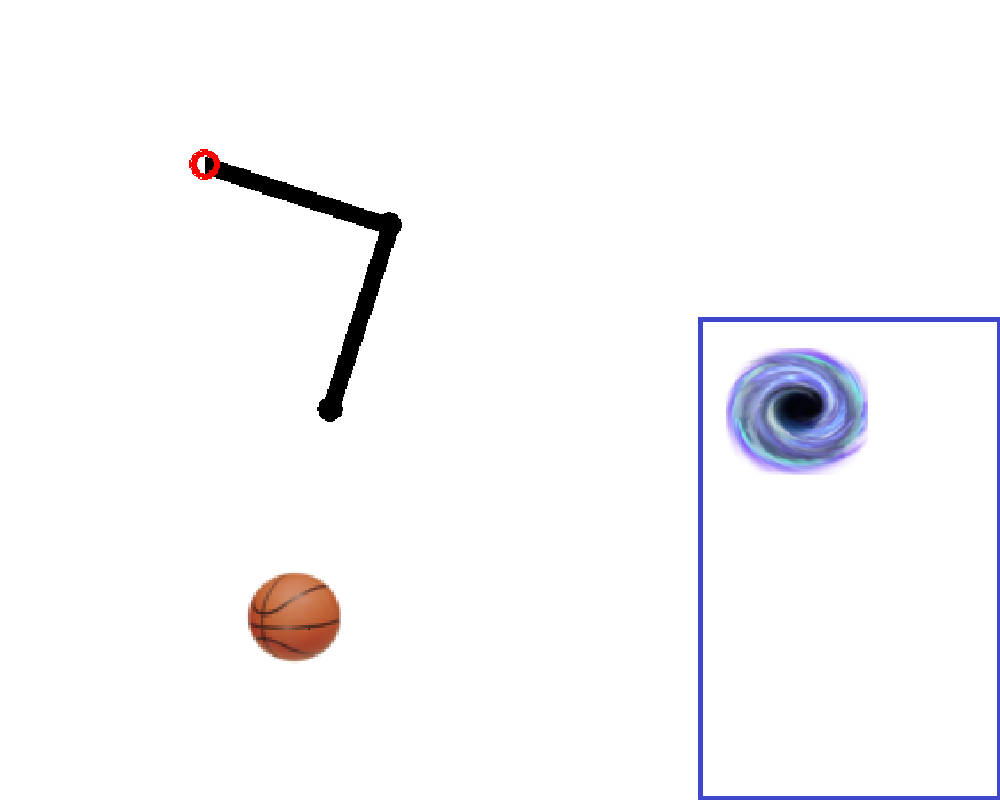}}
		\caption{Robot Environment}
		\label{fig:robot-game}
	\end{subfigure}
	\caption{Environments: (a) The agent needs to pick up the ball and throw it towards the black-hole. (b) The agent needs to pick up the ball and throw it above the wall towards the black-hole. (c) The agent needs to pick the ball with the end-effector by controlling joint velocities and throw it towards the black-hole. The red and black bounding boxes indicate the workspace of the hand and the region of possible target locations, respectively. The objects' dimensions were enlarged for visualization purposes.}
	\label{fig:environments}
\end{figure*}

The original HER algorithm was tested on three environments (OpenAI-Gym), consisting of the following fetch tasks \cite{plappert2018multi, brockman2016openai}: \emph{Push}, \emph{Slide} and \emph{Pick and Place}. The fetch environments include a manipulator and the agent, which controls the end-effector position. The manipulator's state-space is contained within the goal-space, and thus, is incompatible with our condition of having separated goal- and initial-state distributions. We have therefore built three new environments in Python using Pygame \cite{shinners2011pygame}, which are more complex in terms of control than the ones proposed by Open-\\AI:
\begin{enumerate}
	\item \textbf{Hand:} In this task, the hand needs to pick up the ball and throw it at the target (Fig.\ref{fig:hand-game}). 
	\item \textbf{Hand-Wall}: Same as the \textit{Hand} task, but in addition, a wall is placed in-between the agent's workspace and the target. The agent needs to throw the ball above the wall (Fig.\ref{fig:hand-wall-game}).
	\item \textbf{Robot}: In this task, a manipulator needs to pick up the ball and throw it at the target (Fig.\ref{fig:robot-game}). The agent controls the end-effector via the joint velocities, similar to real-world scenarios.
\end{enumerate}
Similar to the OpenAI environments, in both, the Hand and Hand-Wall tasks, the agent controls the end-effector position. However, in all our environments, the agent must learn to throw the ball in the right moment with the right velocity.
For the full description of the environments, see Appendix \ref{appndx:C}.
	
\subsection{Algorithms Performances} \label{Performances}

Training is performed using the DDPG algorithm \cite{lillicrap2015continuous}, in which the actor and the critic were represented using multi-layer perceptrons (MLPs). See Appendix \ref{appndx:D} for more details regarding network architecture and hyperparameters. In order to test the performance of the algorithms, we ran on each environment all four combinations: (\rom{1}) \textbf{HER}, (\rom{2}) \textbf{Filtered-HER}, (\rom{3}) \textbf{HER} with \textbf{IBS} and (\rom{4}) \textbf{Filtered-HER} with \textbf{IBS}. In all algorithms we used prioritized experience replay (PER) \cite{schaul2015prioritized}. The results of the algorithms are evaluated using the following criteria: (\rom{1}) Virtual goal distributions, (\rom{2}) Success rate, (\rom{3}) Distance-to-goal and (\rom{4}) Q-function estimation.
The first criterion analyzes the differences in virtual goal selection for the different algorithms, the second and third evaluate the performances of the agent and the fourth represents the agent's bias.

\subsubsection{Virtual Goal Distributions}
\begin{figure}
	\centering
	\centering
	\includegraphics[width=20em]{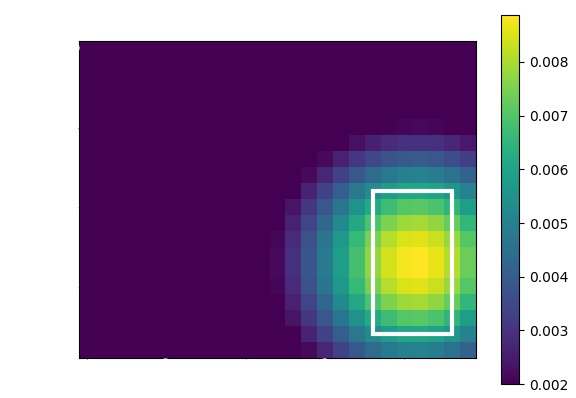}
	\caption{Target Distribution. The target distribution is calculated for $\sigma=0.2$ (screen size is $1\times1 $ in dimensionless units) and minimum value clipped to 0.002. Brighter colors indicate greater ratio of virtual goals.}
	\label{fig:target_dist}
\end{figure}

We compare the virtual goals distributions, resulting from the different algorithms, with the target distribution $q^\ast$ derived in (\ref{eq:optVG}) and visualized in (Fig.\ref{fig:target_dist}). The comparison is performed, both, visually by plotting the distributions, and analytically by using the 
\textit{Kullback-Leibler} (KL) divergence. The KL divergence is defined by 
\bea
K\!L(P||Q)=\sum_{x\in \mathcal{X}}P(x)\,\log\bigg(\frac{P(x)}{Q(x)}\bigg)\eea
and measures the difference between two probability distributions, $P$ and $Q$, which correspond in our case to the target and proposed distributions, respectively.

Table 1 shows the effect of different virtual goal selection strategies and the resulting distributions. The virtual goal distribution generated by \textit{Filtered-HER-IBS} is the closest to the target distribution as indicated by the KL distance in Table 2. As shown in Table 1 \textit{Filtered-HER} reduces dramatically the number of samples on the floor ($y=0$) by removing misleading samples.

\begin{table*}
\begin{minipage}[h]{0.7\textwidth}
	\centering
	\begin{tabular}{m{0em}*5{c}}
		\toprule
		& \textbf{HER} & \textbf{HER-IBS} & \textbf{Filtered-HER} & \textbf{Filtered-HER-IBS}\\
		\midrule
		\vspace*{-1.8cm}\hspace*{-0.5cm}\rotatebox{90}{\textbf{Hand}} & 
		\includegraphics[width=8em]{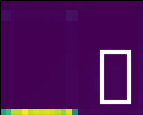} &
		\includegraphics[width=8em]{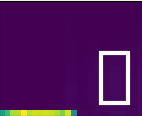} & \includegraphics[width=8em]{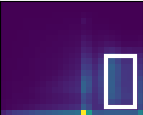} &
		\includegraphics[width=8em]{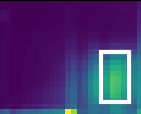} 
		\\
		\cmidrule(lr{1em}){1-5}
		\vspace*{-1.8cm}\hspace*{-0.5cm}\rotatebox{90}{\textbf{Hand-Wall}} & 
		\includegraphics[width=8em]{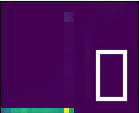} & 
		\includegraphics[width=8em]{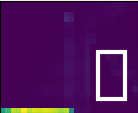} &
		\includegraphics[width=8em]{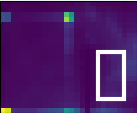} & 
		\includegraphics[width=8em]{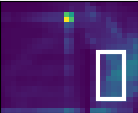}
		\\
		\cmidrule(lr{1em}){1-5}
		\vspace*{-1.8cm}\hspace*{-0.5cm}\rotatebox{90}{\textbf{Robot}} & 
		\includegraphics[width=8em]{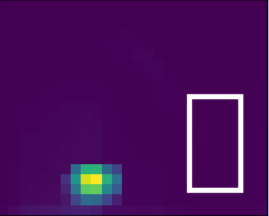} & 
		\includegraphics[width=8em]{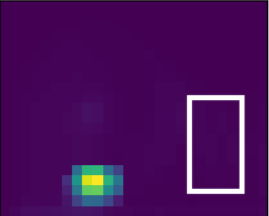} &
		\includegraphics[width=8em]{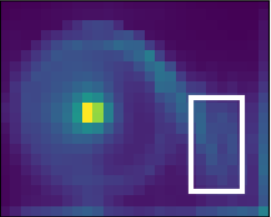} & 
		\includegraphics[width=8em]{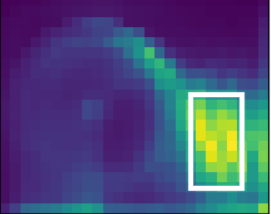}
		\\
		\bottomrule
	\end{tabular}\vspace*{0.1cm}\\	    
	\label{tbl:vg-distribution}
	\hspace*{0.7cm}{\textbf{Table 1} Proposal distributions of virtual goals. Brighter colors indicate greater ratio of virtual goals.}
    % \caption{Proposal distributions of virtual goals. Brighter colors indicate greater ratio of virtual goals.}
    % \label{tbl:vg-distribution}
\end{minipage}
\begin{minipage}[h]{0.1\textwidth}
\hspace*{0.5cm}\includegraphics[width=4em]{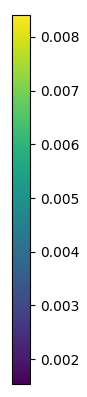}
\end{minipage}
\end{table*}

\bc
\begin{table*}
    \label{tbl:KL-distance}
	\begin{tabular}{ccccc}
		\toprule
		& HER & HER-IBS & Filtered-HER & Filtered-HER-IBS \\
		\cmidrule(lr){2-5}
		\textbf{Hand} &2.0317&1.6623&0.3436&\textbf{0.2272}\\
		\textbf{Hand-Wall} &5.0574&4.6005&0.9606&\textbf{0.5971}\\
		\textbf{Robot} &2.3201&1.9909&0.8609&\textbf{0.3113}\\
		\bottomrule
	\end{tabular}\vspace*{0.1cm}\\
	\hspace*{0.8cm}{\textbf{Table 2} KL Distance}
\end{table*}
\ec

\subsubsection{Success Rate and Distance from Goal}
\begin{figure*}
	\centering
	\begin{subfigure}{.5\textwidth}
		\centering
		\hspace*{-2cm}\includegraphics[width=12cm]{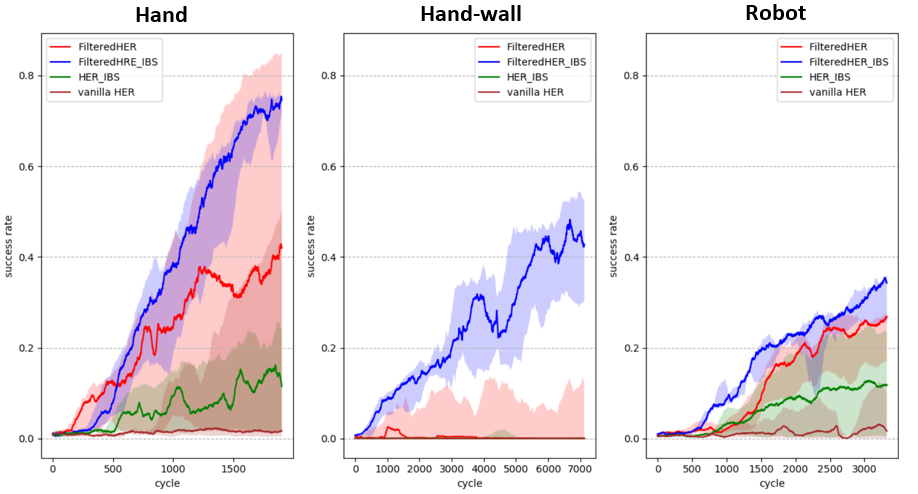}
		\caption{Success rate}
		\label{fig:success_rate}
	\end{subfigure}%
	\\
	\begin{subfigure}{.5\textwidth}
		\centering
		\hspace*{-2cm}\includegraphics[width=12cm]{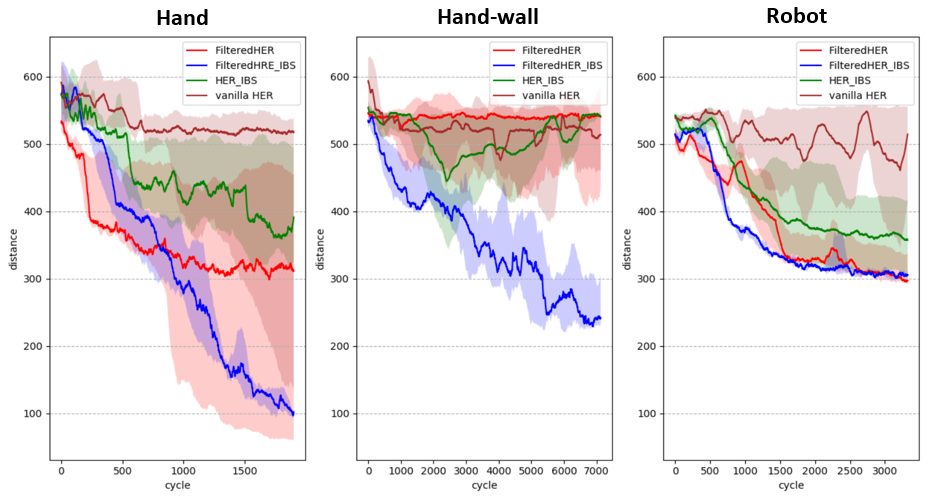}
		\caption{Distance to goal}
		\label{fig:distance_to_goal}
	\end{subfigure}
	\caption{Learning curves for multi-goal manipulation tasks - Hand, Hand-wall, Robot: (a) Success rate. (b) Distance to goal. Results are shown over 15 independent runs. The bold line shows the median and the light area indicates the range between the 33th to 67th percentile. }
	\label{fig:performance}
\end{figure*}

As shown in Fig.\ref{fig:success_rate} and \ref{fig:distance_to_goal}, the vanilla-HER algorithm fails to solve these tasks with nearly zero success rate and almost no improvements in the distance-to-goal measure.
Without using \textit{Filtered-HER}, the agent observes too many misleading samples and fails to learn. Although \textit{Filtered-HER} improved the success rates in all tasks, the performances can be further increased by using IBS for virtual goal selection. Moreover, IBS leads to more robust performances, as indicated by the reduced range of the 33rd to 67th percentile. Performances are mainly affected by two factors: the complexity to affect the achieved goal (e.g., moving the ball) and the complexity to reach the goal (e.g., throwing the ball towards the target). 

Filtered-HER will have more significant impact on tasks, where it is difficult to affect the achieved goal, while IBS will be helpful for tasks, where it is difficult to reach the target. For example, in the \textit{Robot} task, unlike the \textit{Hand} tasks, applying a constant velocity will not result in the hand to get stuck at walls, but rather help to reach almost every possible location in workspace. Therefore, it is simpler to reach the ball in the \textit{Robot} task  than in the \textit{Hand} tasks, and thus, Filtered HER is not as effective for the \textit{Robot} task as it is for the \textit{Hand} tasks. After reaching the ball, the agent tries to throw the ball towards the target. For the \textit{Hand-Wall} task, it is harder to reach the goal, and thus, the agent does not see as many instructive samples. In this case, IBS is especially important as it exploits as much information as possible.

\subsubsection{Estimated Q-value} \label{ss:Estimated Q-value}
\begin{figure*}
    \centering
    \includegraphics[width=12cm]{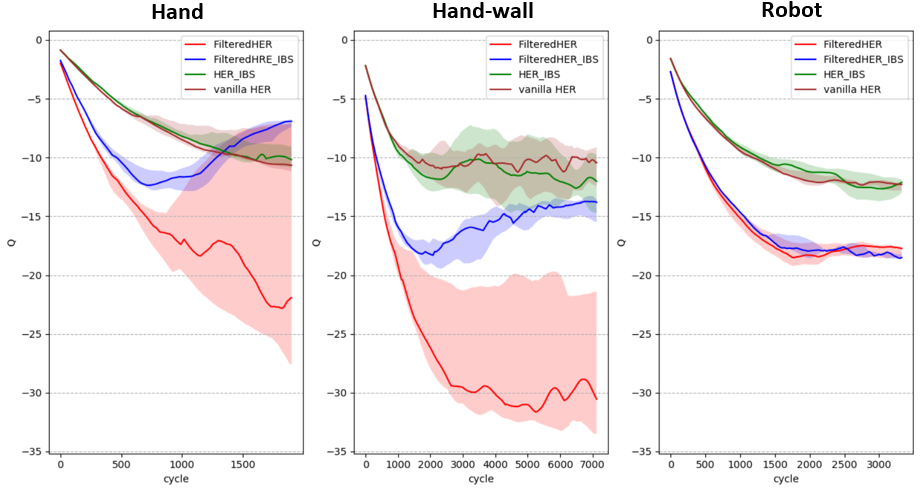}
    \caption{Bias evaluation using the $Q$-values of the initial state and action. Results are shown over 15 independent runs. The bold line shows the median, and the light area indicates the range between the 33rd to 67th percentile.}
    \label{fig:Estimated Q value}
\end{figure*}

Misleading samples lead to optimistic estimates of the action-value function, i.e., the agent overestimates its performance. To demonstrate the bias, Fig.\ref{fig:Estimated Q value} compares the agent's evaluation for the $Q$-value of the initial state and action. Note that the unfiltered versions, although performing poorly as shown in Fig.\ref{fig:success_rate}, led to higher $Q$-value estimates than the better performing filtered versions (except for the \textit{Hand}-task, where the FilteredHER\_IBS performed so much better than the unfiltered versions that its $Q$-value estimate exceeded the biased estimations). Thus, the filter reduces the bias consistently and leads to more accurate evaluation of future returns (Appendix B).

\section{Related Work}
Prioritizing samples over their relevance to the learning has been used in several previous papers, such as  \textit{Prioritized Experience Replay} (PER) \cite{schaul2015prioritized}, \textit{Energy-Based Hindsight Experience Prioritization} (EBP) \cite{zhao2018energy}, \textit{Hindsight Experience Replay With Experience Ranking} (HER+ER) \cite{nguyen2019hindsight} and \textit{Curriculum-guided Hindsight Experience Replay} (CHER) \cite{fang2019curriculum}.
Similar to our algorithm, PER gives higher priority to samples that are unknown to the agent. However, unlike IBS, PER uses the \textit{TD-Error} of the sample to measure the agent's knowledge (i.e., a smaller error implies more acquaintance). PER receives the buffer as a given input set and prioritizes when sampling from it for experience replay. In contrast, IBS prioritizes when building the buffer during experiences. In addtion, unlike IBS, PER only prioritizes over unfamiliar samples and does not take into consideration that some samples might be better towards task completion than others. EBP applies a different prioritization scheme by calculating the amount of (translational and rotational) kinetic energy transferred to the object during an episode. Trajectories associated with a larger kinetic energy transfer are therefore preferred, assuming that the agent can learn more from trajectories in which the object moved significantly. EBR does not differentiate between movement directions and is thus applicable for cases where all directions are equally informative for learning. Similar to PER, EBP receives the buffer as a given input set and prioritizes when sampling from it for experience replay. Since PER and EBP prioritize during experience replay, both methods can be applied with IBS. As in our work, both HER+ER and CHER prioritizes virtual goals, enabling the agent to learn better how to achieve real goals. Yet, there is a fundamental difference between these and our methods. When prioritizing, HER+ER and CHER only considers the similarity between the proposed virtual goals and the desired goal of this specific episode, instead of taking the entire goal distribution into consideration. Therefore, the algorithm may skip virtual goals, which are instructive for other possible goals. Furthermore, when prioritizing, HER+ER only takes the similarity to the real goal into consideration and ignores, whether the agent has already learned a certain region of state space, resulting in an overflow of samples from that location. Although CHER considers the sparsity of virtual goals, a weighting between exploration (sparsity) and exploitation (relevance) must be specified, whereas in our method it is embedded by introducing a single importance measure (the instructiveness), which combines both aspects. In addition, our method takes all previous virtual goal distributions into considerations, when assigning scores to new potential virtual goals. The bias induced by HER has been already recognized in \cite{plappert2018multi}, and a solution was suggested in \textit{Aggressive Rewards to Counter Bias in Hindsight Experience Replay} (ARCHER) \cite{lanka2018archer}. To reduce bias, ARCHER multiplies all virtual rewards by a decreasing factor. Thus, ARCHER does not distinguish between biased and unbiased virtual samples and applies a bias reduction to all virtual samples. In Filtered HER bias-inducing samples are removed a-priori.

\section{Conclusion}
	In this paper we have introduced two novel techniques: an \textit{Instructional-Based Strategy} (IBS) for virtual goal selection and a \textit{Filtered-HER} for the removal of misleading samples. IBS is used for prioritizing more instructive \textit{virtual goals} when collecting experiences while \textit{Filtered-HER} is used to reduce bias that may occur when using HER. Both methods showed significant improvements in performances and sample efficiency when compared to vanilla-HER in the tested environments. Like HER, our methods can be applied using any off-policy RL algorithm, such as DDPG. Moreover, the presented methods can be easily combined with any experience replay prioritization technique as we have demonstrated in our experiments using PER.

\clearpage
\newpage
\onecolumn
\appendix
\section{Spatio-temporal goal distributions}
\label{appndx:A}
In this appendix we consider non-uniform and non-stationary goal  distributions. 

\subsection{Non-uniform goal distributions}
Our approach can be extended to non-uniform goal distributions $g(\bm{x})$. It is sufficient to assume that the goal distribution is provided in terms of samples. We can then represent the goal distribution $g(\bm{x})$ as a Gaussian mixture model (GMM) given by
\bea
g(\bm{x}) &=& \sum_{i=1}^N p_i {\cal{N}} (\bm{x}|\bm{\mu}_i,\bm{\Sigma}_i)\,, \quad \sum_{i=1}^N p_i =1 \,,  \quad 0\le p_i \le 1 \label{A1}
\eea
and use \textit{Expectation Maximization} (EM) to estimate the parameters of the GMM model. After estimating the parameters, the integral (\ref{3}) with Gaussian kernel (\ref{2}) can be performed since the product of two Gaussians is again a (un-normalized) Gaussian. Generally, it is
\bea
\prod_{k=1}^K {\cal{N}}(\bm{x}|\bm{\mu}_k,\bm{\Sigma}_k) &=& Z\cdot {\cal{N}}(\bm{x}|\bm{\mu},\bm{\Sigma})\,, \quad Z = \frac{ | 2\pi {\bm{\Sigma}} |^{\frac{1}{2}} }{ \prod_{k=1}^K |2\pi \bm{\Sigma}_k|^{\frac{1}{2}}} \exp\Bigg[\frac{1}{2} \Bigg( {\bm{\mu}}^T {\bm{\Sigma}}^{-1} {\bm{\mu}} - \sum_{k=1}^K {\bm{\mu}}^T_k \bm{\Sigma}_k^{-1} \bm{\mu}_k  \Bigg)\Bigg]  \label{A2} \eea

with $\bm{\Sigma} = \big( \sum_{k=1}^K \bm{\Sigma}_k^{-1} \big)^{-1}$ and  $\bm{\mu} = \bm{\Sigma}  \big(\sum_{k=1}^K \bm{\Sigma}_k^{-1}\bm{\mu}_k \big)$. 
Using (\ref{A1}), (\ref{A2}), the goal distribution (\ref{3}) can be evaluated leading to
	\bea \label{A3}
	\mu({\tilde{\bm{g}}}|{\sigma})= \int_{\bm{x}}
	k(\tilde{\bm{g}},\bm{x} | \sigma) g(\bm{x}) d\bm{x} =  (2\pi)^{n/2}|\bm{\Sigma}_{\tilde{\bm{g}}}|^{1/2}\int_{\bm{x}}\sum_{i=1}^N p_i {\cal{N}} (\bm{x}|\bm{\mu}_i,\bm{\Sigma}_i)  {\cal{N}} (\bm{x}|\tilde{\bm{g}},\bm{\Sigma}_{\tilde{\bm{g}}}) d\bm{x} = (2\pi)^{n/2}|\bm{\Sigma}_{\tilde{\bm{g}}}|^{1/2}  \sum_{i=1}^N p_i Z_i\,,
	\eea
	where
	\bea
	Z_i  =&= \frac{ | 2\pi {\bm{\Sigma}} |^{\frac{1}{2}} }{ |2\pi \bm{\Sigma}_i|^{\frac{1}{2}} + |2\pi \bm{\Sigma}_{\tilde{\bm{g}}}|^{\frac{1}{2}}   } \exp\Bigg[\frac{1}{2} \Bigg( {\bm{\mu}}^T {\bm{\Sigma}}^{-1} {\bm{\mu}} - {\bm{\mu}}^T_i \bm{\Sigma}_i^{-1} \bm{\mu}_i - \tilde{\bm{g}}^T \bm{\Sigma}_{\tilde{\bm{g}}}^{-1} \tilde{\bm{g}}  \Bigg)\Bigg]
\eea	
with $\bm{\Sigma} = \big(  \bm{\Sigma}_i^{-1} +  \bm{\Sigma}_{\tilde{\bm{g}}}^{-1} \big)^{-1}$ and $\bm{\mu} = \bm{\Sigma}  \big( \bm{\Sigma}_i^{-1}\bm{\mu}_i + \Sigma_{\tilde{\bm{g}}}^{-1}\tilde{\bm{g}} \big)$. \\

%When handling a non uniform distributions, the score cannot be simplify and thus the user must calculate the full integral. However, the resulted integral may be incalculable. If so, 
Alternatively, the integral can be evaluated using \emph{Monte Carlo integration}: Let $\Omega \in \mathbb{R}^n$ and $V=\int_\Omega dx$ be the game's domain. Given $N$ uniform samples, 
$\bm{X}_1,\cdots,\bm{X}_N \sim U(\Omega)$, we can approximate the goal distribution as
\bea \label{eq:MonteCarlo evaluation}
\mu({\tilde{\bm{g}}}|\bm{\sigma}) \approx \tilde{\mu}({\tilde{\bm{g}}}|\bm{\sigma}) \equiv \frac{V}{N}\sum_{i=1}^{N}
(k(\tilde{\bm{g}},\bm{X}_i | \sigma) g(\bm{X}_i)) \,.
\eea
Due to the law of large numbers, it is $\lim_{N\rightarrow\infty} \tilde{\mu}({\tilde{\bm{g}}}|\bm{\sigma})= \mu({\tilde{\bm{g}}}|\bm{\sigma})$. Although the true value is obtained for large $N$, the induced variance may be significant. To reduce variance, importance sampling can be applied. Instead of sampling $\bm{X}_1,\cdots,\bm{X}_N$ from a uniform distribution, one can sample from a chosen distribution $h(x)$. It can be shown that $V\!\!ar(\tilde{\mu}({\tilde{\bm{g}}}|{\sigma}))$ is minimized when sampling $\bm{X}_1,\cdots,\bm{X}_N$ from $h(\bm{x})$ such that $h(\bm{x}) \propto \mu({\tilde{\bm{g}}}|{\sigma})$, which is unknown (see \cite{owen2013importance}).
A good candidate for the sampling distribution $h(\bm{x})$ is the normalized RBF kernel, which is the Gaussian distribution $\mathcal{N}(\bm{x}|\tilde{\bm{g}}, \sigma)$. A second option may be a \textit{Uniform} distribution within the domain of the goal distribution. Yet, the goal distribution's range may be very large or even go to infinity (for example, if the goal distribution is an exponential distribution). For these cases, it is recommended to use the Gaussian distribution for sampling.Although Monte Carlo integration is computationally costly, the algorithm requires to evaluate the integral only once per $\sigma$ value used in training, which can be performed off-line.

\subsection{Non-stationary goal distributions}
Our method was not developed for non-stationary goal distributions, yet in most cases, it should still work fine. As explained in section \ref{sec:implementation}, we change the value of $\sigma$ overtime to stabilize training. Changing the $\sigma$ hyper-parameter over time has a similar effect to changing the goal distribution. As long as the target distribution of virtual goals is updated to the new goal distribution, our algorithms 
are able to adapt. Performance may be affected in extreme cases, where the goal distribution changes drastically. For example, IBS may end up with an almost useless virtual-goal distribution, yet should recover quickly due to its higher prioritization for the most instructive virtual goals.

\newpage
\section{Mathematical motivation for Filtered-HER}
	\label{appndx:B}
In this appendix, we provide a more mathematical motivation for Filtered-HER.\\
In value-function based RL algorithms (e.g., Q-learning, DQN, DDPG, etc.), a $Q$-function $Q(s, a)$ is evaluated for every transition, using the Bellman Equations \cite{sutton2018reinforcement} (we can generalize these algorithms to multi-goal tasks by evaluating $Q(s,a,g)$ \cite{schaul2015universal}). To scale RL algorithms, we use deep neural networks and sample past transitions from a replay buffer \cite{mnih2013playing}. We can formalize the evaluation objective using \emph{Importance Sampling}. If we denote the distribution of real transitions as $\mathcal{P}$, the distribution in the replay buffer as $\mathcal{B}$ and the $(s, a, g)$ tuple as $x$, then the expectation of the $Q$-function is given by

	\bea
	E_{x\sim \mathcal{P}}[Q(x)] = \sum_x \mathcal{P}(x)Q(x)
	= \sum_x \mathcal{B}(x)\left[ \frac{\mathcal{P}(x)}{\mathcal{B}(x)}Q(x)\right] = 
	E_{x\sim \mathcal{B}}\left[ \frac{\mathcal{P}(x)}{\mathcal{B}(x)}Q(x)\right]\,.
	\eea

    In deep reinforcement learning without HER, the replay buffer is sampled directly from the real distribution $\mathcal{P}$, and the buffer size is restricted, thus, mostly containing transitions from recent policies. Therefore, the correction factor ${\mathcal{P}(x)}/{\mathcal{B}(x)}$ is close to $1$, and thus, $E_{x\sim \mathcal{P}}[Q(x)] \approx E_{x\sim \mathcal{B}}[Q(x)]$. However, when using HER, this assumption no longer holds. 
    By adding synthetic virtual transitions to the buffer, the distribution may be changed. Since $\mathcal{P}$ is unknown, we cannot apply the correction factor ${\mathcal{P}(x)}/{\mathcal{B}(x)}$ to the transitions and  our estimation may be biased. In most cases, the effect over the distribution will be minor, and consequently, it will not hinder learning.
    However, as discussed in section \ref{ss:bias in HER} and shown in Table 1, in the absence of a filter, the replay buffer may get overflowed with useless, impossible transitions $\bm{x}^\prime = (s,a,g^\prime)$, where the agent already achieved the goal $g^\prime$ in the previous state $s$ ($r(s,g^\prime) = 0$). Since these transitions always have a non-negative reward, the agent converges to a local optimum between the real and fake transitions. As the agent will never encounter the impossible transitions, while interacting with the environment, the local optimum results in biased, optimistic estimation of the $Q$-function (see section \ref{ss:Estimated Q-value} and Fig.\ref{fig:Estimated Q value}). For these specific transitions, we know that the real distribution is $\mathcal{P}(\bm{x}^\prime) = 0$ since we can never encounter these samples in real episodes.
    Therefore, to reduce bias, we can apply a correction factor equal to $0$ on these transitions, which is equivalent to filtering them out.
    
\newpage
\section{Environments}
	\label{appndx:C}
	
	% Note: in this sample, the section number is hard-coded in. Following
	% proper LaTeX conventions, it should properly be coded as a reference:
	%In this appendix we prove the following theorem from
	%Section~\ref{sec:textree-generalization}:
	In this appendix we describe the environments used for the validation of the algorithms. The environments were built in Python using Pygame. All environments are fully observable, thus, the agents has perfect knowledge of the state of the environment
	\subsection{Hand throwing tasks} \label{ss:hand throwing tasks}
	These tasks include a hand, a ball and a target. The goal in these tasks is to get the ball close enough to the target
	\subsubsection{Hand}
	In this game, the ball is initialized on the ground with probability 0.5 and within the hand otherwise. The agent needs to learn how to pick the ball and throw it towards the target (see Fig.\ref{fig:hand-game}).
	\subsubsection{Hand-Wall}
	This game is like the \textit{Hand} task, but there is also a wall and the agent needs to throw the ball above the wall (see Fig.\ref{fig:hand-wall-game}).
	\subsubsection{Observations}\vspace*{0.3cm}
%	\begin{table}[h!]
%	\caption[Hand\_throw observation]{Hand\_throw observation}
    \bc
		\textbf{Table 3}: Hand\_throw observations \vspace*{0.1cm}\\
		\begin{tabular}{||c | c | c||} 
			\hline
			Num & Observation & Type\\ [0.5ex] 
			\hline\hline
			0 & hand $x$ position & continuous\\ 
			\hline
			1 & hand $y$ position & continuous\\
			\hline
			2 & hand $x$ velocity & continuous\\
			\hline
			3 & hand $y$ velocity & continuous\\
			\hline
			4 & hand state (open/close) & binary\\
			\hline
			5 & ball $x$ position & continuous\\
			\hline
			6 & ball $y$ position & continuous\\
			\hline
			7 & ball $x$ velocity & continuous\\
			\hline
			8 & ball $y$ velocity & continuous\\
			\hline
		\end{tabular}
	\ec
%		\centering
%		\label{tbl:hand-throw observation}
%	\end{table}
	\subsubsection{Actions}\vspace*{0.3cm}
%	\begin{table}[h!]
		\bc
		\textbf{Table 4}: Hand\_throw action \vspace*{0.1cm}\\
		\begin{tabular}{||c | c | c||} 
			\hline
			Num & Action & Type\\ [0.5ex] 
			\hline\hline
			0 & hand $x$ velocity & continuous\\
			\hline
			1 & hand $y$ velocity & continuous\\
			\hline
			2 & hand state (open/close) & binary\\
			\hline
		\end{tabular}
		\ec
%		\label{tbl:hand-throw action}
%	\end{table}
	\subsubsection{Goal}
	%\begin{table}[H]
		\bc
		\textbf{Table 5}: Hand\_throw goal\vspace*{0.1cm}\\
		\begin{tabular}{||c | c | c||} 
			\hline
			Num & Goal & Type\\ [0.5ex] 
			\hline\hline
			0 & black-hole $x$ position & continuous\\
			\hline
			1 & black-hole $x$ position & continuous\\
			\hline
		\end{tabular}
		\ec
		%\label{tbl:hand-throw goal}
	%\end{table}
	\subsubsection{Reward function}
    The reward is binary: $0$ if the target is achieved and $-1$ otherwise:\\
	\begin{center}
		$R(s_{t}) = \begin{cases}
		0,       & ||goal_{pos}-ball_{pos}|| < \epsilon \\
		-1,      & otherwise \\
		\end{cases}$
	\end{center}	
	\newpage
	\subsection{Robot throwing tasks} \label{ss:robot throwing tasks}
	The \emph{Robot} task includes a manipulator, a ball and a target. The goal in this task is to get the ball close enough to the target by controlling the joints velocities of the robotic arm.
	The ball is initialized within the manipulator's reachable area with probability $0.5$ and within the end-effector otherwise. The agent needs to learn how to pick the ball and throw it towards the target (see Fig.\ref{fig:robot-game}).
%	\subsubsection{Observation}
%	\begin{table}[H]
%		\caption[Robot\_throw observation]{Robot\_throw observation}
    \bc
		\textbf{Table 6}: Robot\_throw observation \vspace*{0.1cm}\\
		\begin{tabular}{||c | c | c||} 
			\hline
			Num & Observation & Type\\ [0.5ex] 
			\hline
			0-1* & $\theta$ (joint's angles) & continuous\\ 
			\hline
			2 & end-effector $x$ position & continuous\\
			\hline
			3 & end-effector $y$ position & continuous\\
			\hline
			4-5 & $\dot{\theta}$ (joint's velocity) & continuous\\
			\hline
			6 & end-effector $x$ velocity & continuous\\
			\hline
			7 & end-effector $y$ velocity & continuous\\
			\hline
			8 & end-effector state (open/close) & binary\\
			\hline
			9 & ball $x$ position & continuous\\
			\hline
			10 & ball $y$ position & continuous\\
			\hline
			11 & ball $x$ velocity & continuous\\
			\hline
			12 & ball $y$ velocity & continuous\\
			\hline
		\end{tabular}
    \ec
%		\label{tbl:robot-throw observation}
%	\end{table}
	${}^*$After scaling, $\theta$ is represented by $(\cos \theta, \sin\theta)$.
	\subsubsection{Actions}
%	\begin{table}[H]
    \bc
    \textbf{Table 7}: Robot\_throw action \vspace*{0.1cm}\\
		%\caption[Robot\_throw action]{Robot\_throw action}
		%\centering
		\begin{tabular}{||c | c | c||} 
			\hline
			Num & Action & Type\\ [0.5ex] 
			\hline\hline
			0-1 & $\dot{\theta}$ (joint's velocity) & continuous\\
			\hline
			2 & end-effector state (open/close) & binary\\
			\hline
		\end{tabular}
	\ec
%		\label{tbl:robot-throw action}
%	\end{table}
%	
%	\subsubsection{Goal}
    \bc
	%\begin{table}[H]
	\textbf{Table 8}: Robot\_throw goal \vspace*{0.1cm}\\
		\centering
		\begin{tabular}{||c | c | c||} 
			\hline
			Num & Goal & Type\\ [0.5ex] 
			\hline\hline
			0 & black-hole $x$ position & continuous\\
			\hline
			1 & black-hole $x$ position & continuous\\
			\hline
		\end{tabular}
	\ec
	%	\label{tbl:robot-throw goal}
%	\end{table}
%	
	\subsubsection{Reward function}
	The reward is binary, i.e., $0$ if the target is achieved and $-1$ otherwise:\\
	\begin{center}
		$R(s_{t}) = \begin{cases}
		0,       & ||goal_{pos}-ball_{pos}|| < \epsilon \\
		-1,      & otherwise \\
		\end{cases}$
	\end{center}

\newpage	
 \section{Experiment Details}
	\label{appndx:D}
	In this appendix we provide a description of the experimental details, including networks' architectures and hyper-parameters.
		\subsection{Training algorithm}
	All the training was done using the DDPG algorithm with the following parameters:
%	\begin{center}
\bc
		\begin{tabular}{||c|c||} 
			\hline
			hyper-parameters & value \\ [0.5ex] 
			\hline\hline
			discount factor ($\gamma$) & 0.98\\ 
			\hline
			target-networks smoothing ($\tau$) & 7 \\
			\hline
			buffer size & 1e6 \\
			\hline
			$\epsilon$ initial value & 1 \\
			\hline
			$\epsilon$ decay rate & 0.95 \\
			\hline
			$\epsilon$ final value & 0.05 \\
			\hline
		\end{tabular}
\ec
%	\end{center}
For exploration we used a decaying epsilon-greedy policy:	
	\begin{equation}
	a =
	\begin{cases*}
	a^{*} & with probability $1-\epsilon$ \\
	a^{*}+\mathcal{N}(0,\,I\cdot\sigma)  & with probability $0.8\cdot\epsilon$ \\
	rand(a) & with probability $0.2\cdot\epsilon$
	\end{cases*}\,, \nn
	\end{equation}
	where $\sigma=0.05\cdot action\_range$ and $\epsilon$ decays at the beginning of every epoch.\\
	For experience replay we used \textit{prioritize experience replay} \cite{schaul2015prioritized}.
	\subsection{Neural networks}
	We used the same neural network layout for all the experiments:
	\subsubsection{Actor:}
	\begin{center}
	%\begin{table*}
		\begin{tabular}{||c|c|c|c|c|c||} 
			\hline
			layer & size & type & activation & BN  & additional info\\ [0.5ex] 
			\hline\hline
			input&input dim&Input&relu&No&No \\
			\hline
			hidden 1&64&FC&relu&No&No \\
			\hline
			hidden 2&64&FC&relu&No&No \\
			\hline
			hidden 3&64&FC&relu&No&No \\
			\hline
			output&action dim&FC&tanh&No&No \\
			\hline
		\end{tabular}
		%\end{table*}
	\end{center}
	\begin{center}
		\begin{tabular}{||c|c||} 
			\hline
			hyper-parameter & value \\ [0.5ex] 
			\hline\hline
			learning rate & 0.001 \\
			\hline
			gradient clipping & 3 \\
			\hline
			batch size & 64 \\ 
			\hline
		\end{tabular}
	\end{center}
	
	\subsubsection{Critic:}
	\begin{center}
		\begin{tabular}{||c|c|c|c|c|c||} 
			\hline
			layer & size & type & activation & BN  & additional info\\ [0.5ex] 
			\hline\hline
			input&input dim&Input&relu&Yes&No \\
			\hline
			hidden 1&64&FC&relu&Yes&concat the layer to the action \\
			\hline
			hidden 2&64&FC&relu&Yes&No \\
			\hline
			hidden 3&64&FC&relu&Yes&No \\
			\hline
			output&1&FC&linear&Yes&No \\
			\hline
		\end{tabular}
	\end{center}
	\begin{center}
		\begin{tabular}{||c|c||} 
			\hline
			hyper-parameter & value \\ [0.5ex] 
			\hline\hline
			learning rate & 0.001 \\
			\hline
			gradient clipping & 3 \\
			\hline
			batch size & 64 \\ 
			\hline
		\end{tabular}
	\end{center}
%Appendix sections are coded under \verb+\appendix+.
%
%\verb+\printcredits+ command is used after appendix sections to list 
%author credit taxonomy contribution roles tagged using \verb+\credit+ 
%in frontmatter.
\twocolumn
%\printcredits

%% Loading bibliography style file
%\bibliographystyle{model1-num-names}
% \bibliographystyle{cas-model2-names}
\bibliographystyle{unsrt}

% Loading bibliography database
\bibliography{references}

%%\vskip3pt
%
% \bio{}
%Author biography without author photo.
%Author biography. Author biography. Author biography.
%Author biography. Author biography. Author biography.
%Author biography. Author biography. Author biography.
%Author biography. Author biography. Author biography.
%Author biography. Author biography. Author biography.
%Author biography. Author biography. Author biography.
%Author biography. Author biography. Author biography.
%Author biography. Author biography. Author biography.
%Author biography. Author biography. Author biography.
%\endbio
%
%\bio{figs/pic1}
%Author biography with author photo.
%Author biography. Author biography. Author biography.
%Author biography. Author biography. Author biography.
%Author biography. Author biography. Author biography.
%Author biography. Author biography. Author biography.
%Author biography. Author biography. Author biography.
%Author biography. Author biography. Author biography.
%Author biography. Author biography. Author biography.
%Author biography. Author biography. Author biography.
%Author biography. Author biography. Author biography.
%\endbio
%
%\bio{figs/pic1}
%Author biography with author photo.
%Author biography. Author biography. Author biography.
%Author biography. Author biography. Author biography.
%Author biography. Author biography. Author biography.
%Author biography. Author biography. Author biography.
%\endbio

\end{document}